\documentclass[conference]{IEEEtran}
\IEEEoverridecommandlockouts

\usepackage{cite}
\usepackage{amsmath,amssymb,amsfonts}
\usepackage{algorithmic}
\usepackage{graphicx}
\usepackage{textcomp}
\usepackage{xcolor}
\usepackage{url}
\def\BibTeX{{\rm B\kern-.05em{\sc i\kern-.025em b}\kern-.08em
    T\kern-.1667em\lower.7ex\hbox{E}\kern-.125emX}}
\begin{document}

\title{MCP-AI: Protocol-Driven Intelligence Framework for Autonomous Reasoning in Healthcare\\


}

\author{
\author{\IEEEauthorblockN{Zag ElSayed, Senior Member IEEE}
\IEEEauthorblockA{\textit{School of Information Technology} \\
\textit{University of Cincinnati}\\
Ohio, USA 
}
\and
\IEEEauthorblockN{Craig Erickson, Ernest Pedapati}
\IEEEauthorblockA{\textit{Division of Child Neurology and Adolescent Psychiatry} \\
\textit{Cincinnati Children’s Hospital Medical Center}\\
Ohio, USA
}
}
}

\maketitle

\begin{abstract}
Healthcare AI systems have historically faced challenges in merging contextual reasoning, long-term state management, and human-verifiable workflows into a cohesive framework. This paper introduces a completely innovative architecture and concept: combining the Model Context Protocol (MCP) with a specific clinical application, known as MCP-AI. This integration allows intelligent agents to reason over extended periods, collaborate securely, and adhere to authentic clinical logic, representing a significant shift away from traditional Clinical Decision Support Systems (CDSS) and prompt-based Large Language Models (LLMs). As healthcare systems become more complex, the need for autonomous, context-aware clinical reasoning frameworks has become urgent. We present MCP-AI, a novel architecture for explainable medical decision-making built upon the Model Context Protocol (MCP) a modular, executable specification for orchestrating generative and descriptive AI agents in real-time workflows. Each MCP file captures clinical objectives, patient context, reasoning state, and task logic, forming a reusable and auditable memory object. Unlike conventional CDSS or stateless prompt-based AI systems, MCP-AI supports adaptive, longitudinal, and collaborative reasoning across care settings. MCP-AI is validated through two use cases: (1) diagnostic modeling of Fragile X Syndrome with comorbid depression, and (2) remote coordination for Type 2 Diabetes and hypertension. In either scenario, the protocol facilitates physician-in-the-loop validation, streamlines clinical processes, and guarantees secure transitions of AI responsibilities between healthcare providers. The system connects with HL7/FHIR interfaces and adheres to regulatory standards, such as HIPAA and FDA SaMD guidelines. MCP-AI provides a scalable basis for interpretable, composable, and safety-oriented AI within upcoming clinical environments.
\end{abstract}

\begin{IEEEkeywords}
MCP, Clinical AI, Generative AI, Descriptive AI, Decision Support Systems, Healthcare, Informatics, Model Context Protocol, Autonomous Reasoning.
\end{IEEEkeywords}

\section{Introduction}
The use of artificial intelligence (AI) in the healthcare sector has progressed swiftly. Nevertheless, current tools tend to be limited in their capabilities and frequently lack a connection to practical clinical decision-making.~\cite{dave2023chatgpt}~\cite{haupt2023ai}. Traditional clinical decision support systems (CDSS) rely on static rules and ontologies~\cite{fda_samd_eval2017}~\cite{hagerman2017fragilex}. At the same time, modern generative models, including large language models (LLMs), produce plausible narratives without internal memory, state persistence, or task logic~\cite{singhal2023large}. These systems lack the structure and interpretability necessary for high-stakes medical decision-making, and they are unable to adapt to longitudinal workflows or cross-specialty handoffs.
Simultaneously, clinicians face increasing cognitive and procedural complexity: synthesizing vast volumes of heterogeneous data, updating care plans dynamically, and reasoning across multidisciplinary teams. This challenge is especially acute in contexts such as mental health, chronic care, and rare disease diagnosis, where decisions are temporal, personalized, and uncertainty-laden~\cite{harish2021artificial}~\cite{mckeever2017review}. To address these gaps, AI systems must evolve from static predictors to contextual collaborators.

We introduce MCP-AI, a novel architecture for autonomous clinical reasoning based on the Model Context Protocol (MCP)~\cite{hou2025model}. MCP is a structured, version-controlled file format that captures patient state, clinical objectives, module orchestration logic, and reasoning history~\cite{dave2023chatgpt}~\cite{haupt2023ai}~\cite{singhal2023large}. It enables multi-agent collaboration between generative AI (e.g., for summarization and planning) and descriptive AI (e.g., for rule validation and scoring), all within a persistent, auditable reasoning context~\cite{sutton2020overview}.

Unlike stateless prompts or siloed pipelines, MCP-AI supports real-time task orchestration, modular reasoning, and physician-in-the-loop decision-making~\cite{musen2021clinical}~\cite{berner2007clinical}~\cite{belle2013biomedical}. It operates as a cognitive middleware layer that bridges AI agents with EHR systems, laboratory services, and verification interfaces. This approach empowers machines not only to respond, but to reason and adapt~\cite{wang2019ai}.

In this paper, we describe the design and implementation of MCP-AI and demonstrate its utility through two representative use cases: early-stage neurodevelopmental diagnostics for Fragile X Syndrome with depression, and chronic disease coordination for diabetes and hypertension. These simulations highlight the system’s ability to maintain clinical context across transitions, automate protocol execution, and deliver explainable, traceable medical AI that is aligned with healthcare regulations~\cite{halpern2017reasoning}~\cite{harish2021artificial}~\cite{alli2024potential}.

\section{Background}
The MCP-AI framework is a pioneering architectural approach that merges the general-purpose Model Context Protocol (MCP) with specialized mechanisms for healthcare reasoning. This integration creates a unique cognitive infrastructure designed specifically to connect artificial intelligence with clinical practice. In contrast to earlier frameworks, MCP-AI establishes a systematic, verifiable reasoning environment where both generative and descriptive AI modules work together under protocol oversight, incorporating physician engagement and regulatory protections within the system.

\subsection{Technical Background}
Modern AI applications in healthcare generally exhibit two opposing characteristics: well-organized yet inflexible expert systems, and adaptable but unclear generative models~\cite{sidey2019machine}. Traditional CDSS rely on rule-based engines and curated ontologies such as SNOMED CT and ICD-10~\cite{fda_samd_eval2017}~\cite{hagerman2002fragile}, which provide traceability but lack scalability and adaptability across diverse patient contexts. In contrast, large language models (LLMs) such as GPT-4 and Med-PaLM have demonstrated impressive capabilities in summarizing clinical text and simulating diagnostic dialogue~\cite{raji2023concrete}~\cite{lewandowski2023transforming}, but operate as stateless functions with limited ability to maintain context, explain decisions, or align with regulatory frameworks~\cite{bubeck2023sparks}.

The attempt to combine symbolic and neural methods has produced hybrid systems with some success. However, few solutions provide a comprehensive framework for managing collaboration among multiple agents over time. The Model Context Protocol (MCP) aims to address this challenge by encapsulating not only data and intentions, but also reasoning states, execution histories, and validation mechanisms. It incorporates ideas from microservice architecture, knowledge graphs, and agent-based coordination, facilitating context-aware management across AI modules~\cite{sutton2020overview}~\cite{ochoa2023context}.

\subsection{Clinical Background}
In clinical environments, decision-making is often distributed across providers, departments, and time~\cite{desroches2020views}. Chronic diseases, such as Type 2 Diabetes, require sustained engagement, adaptation to sensor data, and multidisciplinary care coordination~\cite{hooker2022workforce}~\cite{oslock2022contemporary}. Psychiatric and neurodevelopmental disorders such as Fragile X Syndrome add layers of diagnostic uncertainty, requiring longitudinal tracking of behavioral, genetic, and neurophysiological indicators~\cite{protic2022fragile}~\cite{hagerman2002fragile}~\cite{mckeever2017review}.

Most current clinical tools lack the capacity to preserve reasoning continuity across handoffs, track task-level memory, or synthesize multi-modal data into interpretable recommendations~\cite{thornicroft2017undertreatment}. As a result, physicians frequently face gaps in context that compromise care quality, increase the risk of errors, and exacerbate provider fatigue. MCP-AI was designed to mitigate these challenges by providing a shared, persistent reasoning layer that evolves alongside the patient journey~\cite{thilina2020economic}~\cite{zhang2024change}. By bridging generative insights with descriptive guardrails, the system enables AI to participate in, but not overtake, the clinical decision process~\cite{singhal2023large}~\cite{brodsky2025generative}~\cite{rao2025multimodal}.

\section{System Architecture}
\begin{figure}[htbp]
\centerline{\includegraphics[width =\linewidth]{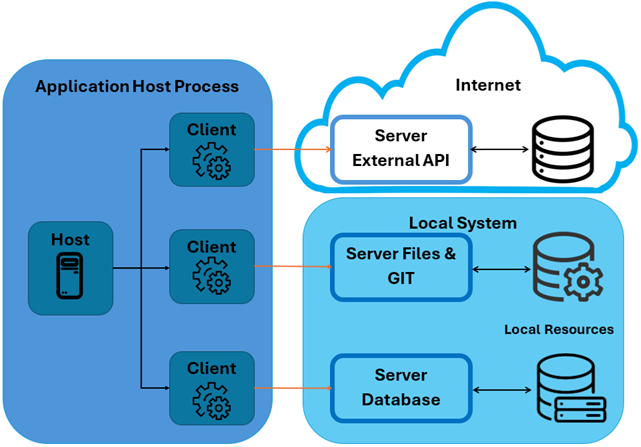}}
\caption{A typical MCP content aware bridge block diagram}
\label{fig2}
\end{figure}

Before delving into the layered structure of MCP-AI, we must first clarify the overall concept of the Model Context Protocol (MCP). MCP is a structured, version-controlled file-based interface that aims to capture task context, execution logic, model orchestration, and real-time decision metadata. Initially developed to streamline communication between intelligent agents and their external environments, MCP offers a standardized framework for the integration of AI modules, APIs, procedural logic, and human input into a traceable and modular workflow. An MCP file generally encompasses patient information, clinical goals, diagnostic hypotheses, execution procedures, fallback scenarios, and confidence annotations, functioning as both a set of computational instructions and a clinical audit trail, shown in Fig.\ref{fig2}.

\section{System Architecture}
The MCP-AI framework represents a groundbreaking change in the fields of engineering design and clinical informatics. Essentially, it connects two previously isolated areas: symbolic protocol-driven task planning and generative deep learning for reasoning in context. From an engineering viewpoint, MCP-AI is crafted as a distributed, interpretable, and composable architecture, allowing for scalability across edge systems, cloud solutions, and embedded environments. It integrates smoothly with HL7/FHIR APIs~\cite{sreejith2023smart}, facilitates the modular enhancement of individual AI components, and ensures state traceability and error recovery through its protocol logic. It establishes the foundation for the development of truly intelligent, compliant, and continuously evolving clinical systems. MCP-AI transcends simple task optimization, it revolutionizes the manner in which AI understands, structures, and participates in care delivery, utilizing traceable and interpretable reasoning logic.

From a systems engineering perspective, this framework enables seamless interaction among microservices, facilitates asynchronous operations, and ensures fail-safe triggers, making it suitable for critical applications such as ICU automation, telemonitoring, and precision diagnostics. The architecture's layers are designed to be modular, enabling hospitals to customize deployments for particular service lines while preserving the integrity of central orchestration.

From a clinical perspective, MCP-AI is groundbreaking as it captures the time-based and procedural thought processes that clinicians use subconsciously, producing a machine-readable format of human medical reasoning. This advancement allows AI systems to transcend mere isolated predictions and engage effectively within actual workflows, including chronic care management, interdisciplinary transitions, and psychiatric decision-making support~\cite{bouchez2024interprofessional}. The architecture is capable of capturing and replicating clinical intent, reasoning justifications, and procedural progression, which are crucial for auditability, safety, and trust in high-stakes environments, particularly because it captures the sequence of medical reasoning as it evolves, something no current AI system can reliably do. For instance, traditional systems may suggest a diagnosis, but they do not retain the logic chain or contextual flags that influenced that suggestion. MCP-AI’s record of clinical intent, data uncertainty, and task progress enables robust oversight and post-hoc explainability.

This framework is crucial for ensuring cognitive coherence when moving between providers, which frequently results in care fragmentation and medical errors. By embedding reasoning pathways and intended outcomes into a durable, easily searchable format, MCP-AI supports seamless transitions among doctors, departments, and healthcare institutions.

In the healthcare-specific implementation, proposed MCP-AI extends MCP with domain-specific capabilities, including HL7/FHIR integration, clinical safety constraints, and physician-in-the-loop verification~\cite{10.1007/978-3-031-91760-8_7}. MCP-AI is designed as a five-layer modular framework that orchestrates autonomous clinical reasoning in a dynamic, traceable, and context-sensitive manner. Each layer operates as an independent but interoperable component in the broader medical AI workflow, supporting clinical decision-making from data ingestion to human verification and system output.

\subsection{Input and Perception Layer}
The system begins by ingesting structured and unstructured clinical data. This includes EHR records, sensor inputs (e.g., EEG, wearable data), patient self-reports, and clinician annotations. The data is semantically normalized and encapsulated in a versioned MCP file, which becomes the operational blueprint for all downstream reasoning and task execution.

\subsection{MCP Engine}
The MCP engine serves as the central hub for control and coordination. It analyzes the MCP file, understands clinical goals, organizes task execution among AI agents, and keeps track of a timeline of activities. This engine ensures consistency, allows for auditing, and supports adaptability during runtime.

\subsection{AI Reasoning Modules}
Two classes of reasoning modules interact with the MCP engine:
\begin{enumerate}
    \item \textbf{Generative AI Modules} (e.g., large language models) generate narrative diagnoses, summaries, and care plans.
    \item \textbf{Descriptive AI Modules} (e.g., probabilistic rule engines, knowledge graphs) validate outputs against clinical guidelines and risk models.
\end{enumerate}
These modules incorporate results into the MCP file, including metadata like confidence scores and explanatory notes.

\subsection{Task and Procedure Agents}
These agents transform validated decisions into clinical actions. For example, Lab Order Agents trigger lab requests via HL7/FHIR APIs, Follow-Up Agents generate alerts and scheduling cues, and Handoff Agents prepare structured summaries for cross-provider transitions.

\subsection{Verification Module and Physician Interface}
\begin{figure}[htbp]
\centerline{\includegraphics[width =\linewidth]{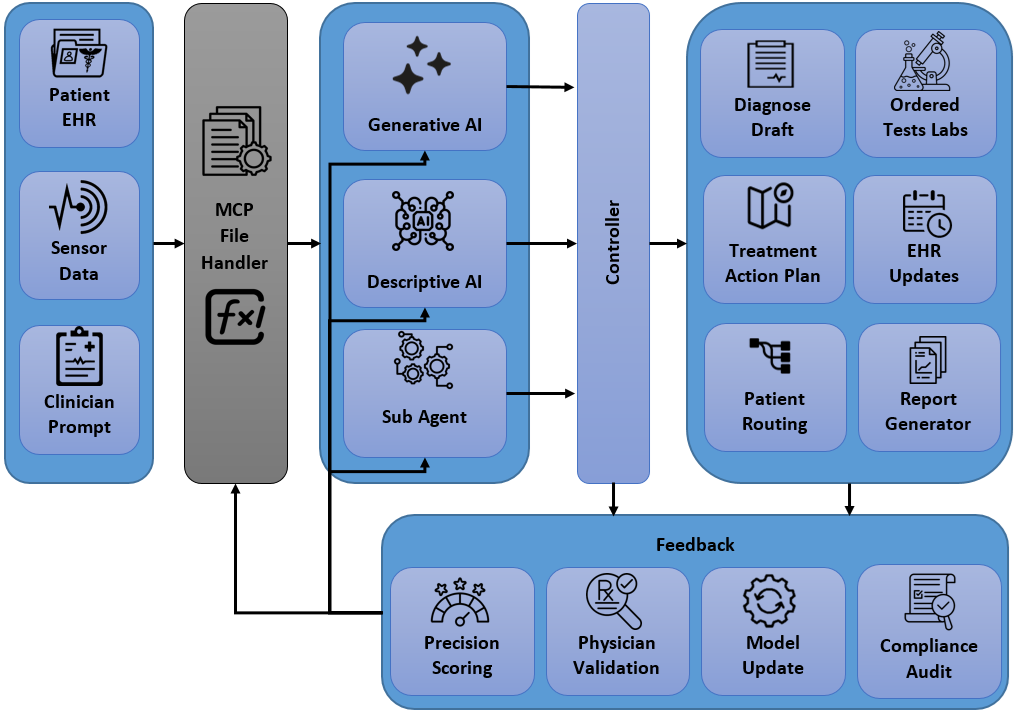}}
\caption{The proposed MCP-AI Healthcare Architecture}
\label{fig1}
\end{figure}

Human verification occurs through an interactive dashboard integrated with clinical workflows. Providers can review suggested plans, simulate alternative actions, and modify or approve execution. This preserves physician oversight and aligns with safety and regulatory requirements. The whole system architecture and data flow are shown in Fig.\ref{fig1}.

\section{Use Case Demonstrations}
Every use case presented here not only highlights the system’s capabilities but also emphasizes its essential position in transforming the future of healthcare AI. MCP-AI offers a fresh perspective on how clinical reasoning, care delivery, and physician oversight are harmoniously integrated. By incorporating logic, memory, and procedural traceability into every scenario, the architecture advances medical AI from a passive decision-support tool to an engaging, interpretive engine of clinical continuity.

The novelty of MCP-AI is most clearly demonstrated through practical use cases. By modeling the thinking process and procedural context in real-time, MCP-AI enables a fundamentally new form of AI-clinician interaction. The following simulations are not mere applications, they represent a new class of AI-driven healthcare workflows, made possible only by this architecture.

To showcase the effectiveness and flexibility of MCP-AI in practical clinical environments, we provide two representative simulation cases. These examples highlight the system's ability to handle intricate diagnostic assessments, carry out care procedures, and maintain continuity over time during clinical transitions.

\subsection{Case: Fragile X Syndrome with Comorbid Depression}
Comorbid depression is defined as having a depressive disorder in addition to another health issue, which is frequently a persistent physical ailment or another mental health disorder~\cite{gold2020comorbid}. The coexistence of these conditions can significantly impact an individual's overall health, quality of life, and effectiveness of treatment outcomes.
Fragile X syndrome is the most prevalent inherited condition that leads to intellectual disability and autism spectrum disorders~\cite{hagerman2002fragile}. It arises from a mutation in the FMR1 gene, resulting in a lack of the FMRP protein, which is essential for proper brain development~\cite{tau2010normal}. This lack can cause a variety of symptoms, such as intellectual disability, delays in development, and issues related to behavior and emotions.
This scenario highlights MCP-AI’s ability to process rare, multimodal diagnostic challenges involving neurodevelopmental and psychiatric factors. It reflects the architecture’s potential to personalize care for vulnerable populations, particularly when dealing with overlapping syndromic, behavioral, and cognitive features.

This situation showcases MCP-AI's capability to handle uncommon, multimodal diagnostic issues that include neurodevelopmental and psychiatric elements. It illustrates the system's potential to tailor treatment for at-risk groups, especially when addressing interconnected syndromic, behavioral, and cognitive characteristics.

A 13-year-old patient presents with behavioral disturbances, developmental delay, and academic difficulties. MCP-AI initializes a new MCP file (e.g., \texttt{MCP-FXS-013}) that incorporates caregiver interviews, behavioral observation notes, EEG waveforms, educational records, and structured data from previous outpatient visits.

The generative module analyzes temporal patterns in these inputs, constructs a narrative hypothesis involving Fragile X Syndrome and depressive symptoms, and suggests a provisional diagnostic tree with sequential decision checkpoints. It generates a structured request for FMR1 gene testing, automated referral to a clinical geneticist, and an integrated behavioral health evaluation. In parallel, the descriptive AI module uses DSM-V and ICD-10 criteria~\cite{smith2018validation} to validate the likelihood of co-occurring diagnoses, while also checking for conflicts, red flags, and missing data in the diagnostic chain.

The MCP protocol logs each of these recommendations, their justifications, confidence intervals, and which AI module contributed to each step. Once reviewed and approved by the pediatric neurologist, the system executes validated orders and creates future task hooks (e.g., three-week follow-up EEG, educational intervention referrals, and pharmacological consult for depressive symptoms).

\begin{figure}[htbp]
\centerline{\includegraphics[width =\linewidth]{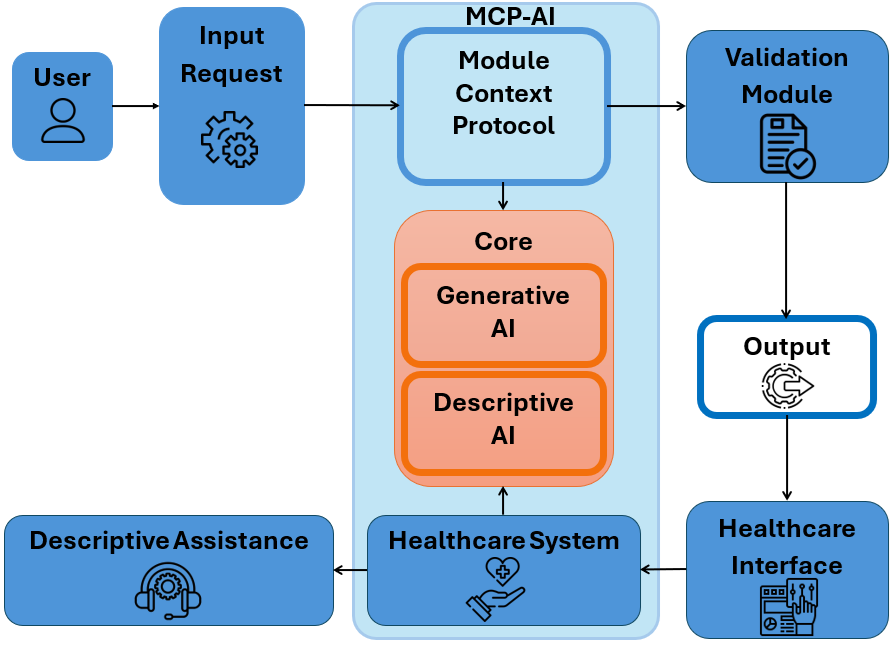}}
\caption{Autonomous clinical reasoning: Fragile X syndrome case}
\label{fig3}
\end{figure}

This case underscores the unique power of MCP-AI to handle diagnostic complexity, manage rare disease pathways, and orchestrate multidisciplinary logic while maintaining clinician alignment and auditable oversight. The holistic operation flow block diagram is shown in Fig.\ref{fig3}.

\subsection{Case: Chronic Care Coordination for Diabetes and Hypertension}
This use case demonstrates MCP-AI’s capacity for continuous, adaptive care planning in common but highly variable chronic conditions. Its ability to manage asynchronous care signals and hand off reasoning logic between human clinicians ensures continuity and mitigates fragmentation, particularly valuable in aging or resource-limited populations.

A 58-year-old patient enrolled in a remote chronic care program presents with glucose variability and medication adherence issues. MCP-AI initializes a new instance (\texttt{MCP-CHRONIC-225}) and aggregates data from wearable devices, glucometer logs, recent A1C trends, and social determinants such as housing instability and food access. The generative AI module drafts a contextual care adjustment including dietary counseling, a titration of metformin, and the potential introduction of an SGLT2 inhibitor~\cite{vallon2015mechanisms}.

Descriptive agents validate the plan against American Diabetes Association (ADA) guidelines~\cite{mckeever2017review}~\cite{thornicroft2017undertreatment}, flag contraindications due to renal function, and compute behavioral adherence probability scores using patient engagement history~\cite{graffigna2015measuring}. The Handoff Agent prepares a concise yet comprehensive reasoning summary embedded in the MCP, which includes rationale for recent changes, alerts for missed labs, and upcoming follow-up tasks.

\begin{figure}[htbp]
\centerline{\includegraphics[width =\linewidth]{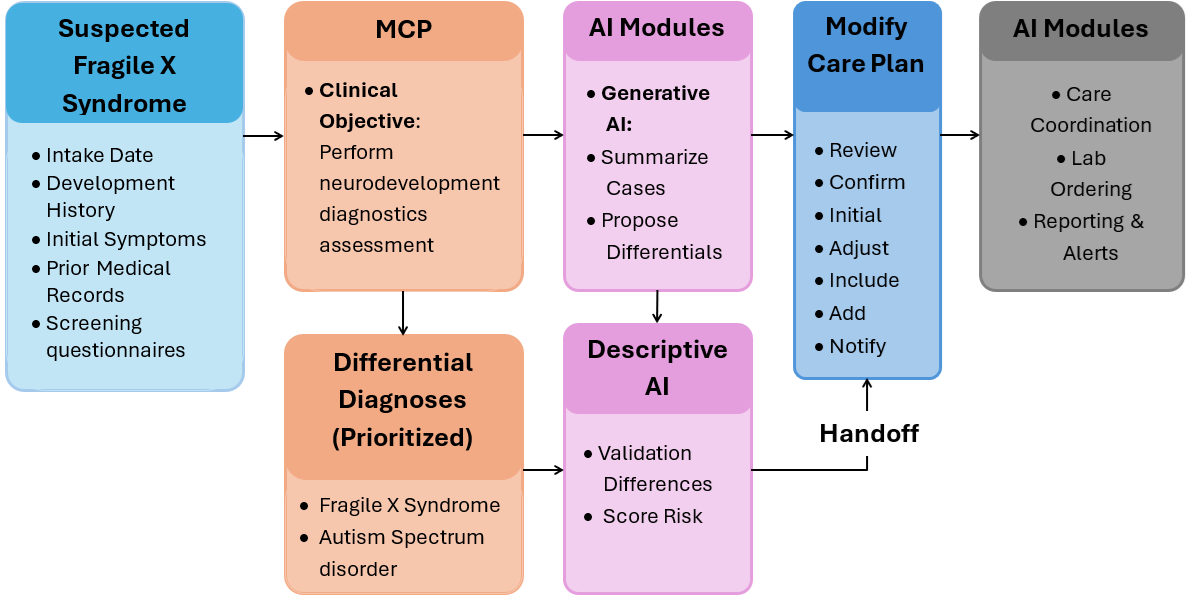}}
\caption{Handoff in chronic care coordination with MCP-AI}
\label{fig4}
\end{figure}

MCP-AI also creates a projected care pathway based on simulated trajectories for glucose and blood pressure levels, which the incoming provider can adjust based on newly surfaced data. Through this mechanism, MCP-AI not only automates and validates decisions but maintains longitudinal understanding of the patient’s evolving clinical state, enabling more consistent, patient-centered, and anticipatory care~\cite{tranfaglia2011psychiatric}, demonstrating how MCP-AI supports structured, explainable decision-making in heterogeneous clinical domains while enabling seamless cross-provider collaboration and compliance with regulatory workflows. The Autonomous clinical reasoning block diagram of the case study is shown in Fig.\ref{fig4}.

\section{Evaluation and Safety Considerations}
To ensure safe, effective, and transparent deployment of MCP-AI in clinical settings, we evaluated the system across several critical dimensions: reasoning accuracy, interpretability, workflow integration, auditability, and regulatory compliance. Each component of MCP-AI was engineered with clinical safety in mind and is designed to function within the constraints of existing healthcare infrastructure.

\subsection{Clinical Accuracy and Verification}
MCP-AI utilizes a tiered validation approach that merges generative recommendations with rule-based verification. Prior to confirming any actions, the system records the reasoning steps, allocates confidence scores, and allows for verification by a physician. This dual-layer protection reduces the risk of errors and facilitates human supervision during crucial decision-making moments.

\subsection{Explainability and Traceability}
Unlike black-box neural systems, MCP-AI logs every step of reasoning, including what was inferred, what data was referenced, which modules were used, and which clinical thresholds were applied. These steps are encoded within the MCP file, allowing for post-hoc inspection and real-time auditability. This traceability is essential for patient safety, clinical trust, and legal defensibility.

\subsection{Workflow Integration and Continuity}
MCP-AI is architected for seamless integration with existing HL7/FHIR-compatible systems and clinical information systems. By generating structured outputs compatible with hospital EMRs, lab APIs, and patient portals, the framework minimizes disruption to existing workflows and ensures longitudinal continuity across multiple care episodes.

\subsection{Regulatory Alignment}
The architecture aligns with key principles of FDA Software as a Medical Device (SaMD), including the ability to track changes in AI behavior over time, support version control, and provide transparent documentation of decision-making processes~\cite{wang2019ai}~\cite{us2021artificial}. The MCP structure supports compliance with HIPAA privacy mandates and enables audit logging suitable for both internal quality review and external regulatory inquiry.

\section{Prospective Scalability and Adaptation}
MCP-AI's modular protocol framework allows for local customization while maintaining architectural uniformity. It is designed to be scalable across inpatient, outpatient, telehealth, and emergency environments, offering tailored task workflows for various departments while centralizing reasoning and safety verification protocols. As new modules, such as imaging agents or genomics interpreters, are introduced, they are integrated through standardized MCP extensions.
Collectively, these characteristics establish MCP-AI as a clinically feasible and future-oriented AI architecture, effectively balancing innovation with accountability, autonomy with regulation, and intelligence with reliability.

\section{Conclusion and Future Work}
MCP-AI represents a pivotal advancement at the crossroads of artificial intelligence and clinical medicine. It is more than just a refinement of existing systems; it signifies a fundamental change in the architecture of intelligent healthcare. By establishing a cohesive, traceable, and adaptable framework that combines generative and descriptive AI with structured protocol logic, MCP-AI reimagines the possibilities in automated medical reasoning. This initiative showcases the inaugural implementation of a cognitive infrastructure where AI goes beyond merely suggesting outcomes; it actively emulates clinical reasoning, retains memory over time, rationalizes its decisions, and collaborates with healthcare professionals in real-time. The repercussions extend well beyond mere technical innovation: MCP-AI has the potential to transform the way healthcare is administered, documented, and regulated. Its capacity to facilitate prolonged reasoning, streamline complex workflows, and ensure clinical oversight positions it as a vital driver of global digital health evolution.
From academic hospitals in high-resource nations to remote clinics in underserved regions, MCP-AI provides a scalable, interpretable, and safety-aligned platform for equitable, intelligent care. This architectural framework establishes the foundation for a new age of AI-assisted healthcare, one in which machines and healthcare professionals work together not through abstraction, but via a common language of organized reasoning, inherent trust, and synchronized decision-making.

\textbf{Future Work}
The authors intend to incorporate real-time biosignal feedback mechanisms (such as EEG and ECG), broaden the task agent libraries to encompass modules for rehabilitation and preventive care, and facilitate distributed learning while ensuring privacy in model updates. Additionally, initiatives are being taken to evaluate MCP-AI against current AI/ML-based diagnostic instruments in prospective clinical trials. These initiatives are designed to transform MCP-AI into a digital reasoning platform that is fully aligned with regulatory standards and clinically validated, aiming to establish a new benchmark for the integration of AI within the healthcare sector.

\bibliographystyle{ieeetr}
\bibliography{references}

\end{document}